\documentclass[sigconf]{acmart}
\AtBeginDocument{%
  }

\copyrightyear{2025}
\acmYear{2025}
\setcopyright{cc}
\setcctype{by-nc-sa}
\acmConference[MM '25]{Proceedings of the 33rd ACM International Conference on Multimedia}{October 27--31, 2025}{Dublin, Ireland}
\acmBooktitle{Proceedings of the 33rd ACM International Conference on Multimedia (MM '25), October 27--31, 2025, Dublin, Ireland}\acmDOI{10.1145/3746027.3755286}
\acmISBN{979-8-4007-2035-2/2025/10}

\usepackage{multirow}
\usepackage{booktabs}

\begin{document}

\title{Disentangle Identity, Cooperate Emotion: Correlation-Aware Emotional Talking Portrait Generation}

\author{Weipeng Tan}
\authornote{All authors contributed equally to this research. This work was done when Weipeng Tan was an intern at Tencent Youtu Lab. Chuming Lin is the corresponding author.}
\email{wptan23@m.fudan.edu.cn}
\orcid{0009-0006-0165-9052}
\affiliation{%
  \institution{Fudan University}
  \city{Shanghai}
  \country{China}
}

\author{Chuming Lin}
\authornotemark[1]
\email{chuminglin@tencent.com}
\orcid{0009-0001-5882-9458}
\affiliation{%
  \institution{Tencent, YouTu Lab}
  \city{Shanghai}
  \country{China}
}

\author{Chengming Xu}
\authornotemark[1]
\email{chengmingxu@tencent.com}
\orcid{0000-0003-3891-2227}
\affiliation{%
  \institution{Tencent, YouTu Lab}
  \city{Shanghai}
  \country{China}
}

\author{FeiFan Xu}
\email{feifanxu@tencent.com}
\orcid{0009-0008-2256-0179}
\affiliation{%
  \institution{Tencent, YouTu Lab}
  \city{Shanghai}
  \country{China}
}

\author{Xiaobin Hu}
\email{xiaobinhu@tencent.com}
\orcid{0000-0002-5764-3096}
\affiliation{%
  \institution{Tencent, YouTu Lab}
  \city{Shanghai}
  \country{China}
}

\author{Xiaozhong Ji}
\email{shawn_ji@163.com}
\orcid{0009-0000-1044-5853}
\affiliation{%
  \institution{Tencent, YouTu Lab}
  \city{Shanghai}
  \country{China}
}

\author{Junwei Zhu}
\email{junweizhu@tencent.com}
\orcid{0000-0002-5407-5150}
\affiliation{%
  \institution{Tencent, YouTu Lab}
  \city{Shanghai}
  \country{China}
}

\author{Chengjie Wang}
\email{jasoncjwang@tencent.com}
\orcid{0000-0003-4216-8090}
\affiliation{%
  \institution{Tencent, YouTu Lab}
  \city{Shanghai}
  \country{China}
}

\author{Yanwei Fu}
\email{yanweifu@fudan.edu.cn}
\orcid{0000-0002-6595-6893}
\affiliation{%
  \institution{Fudan University}
  \city{Shanghai}
  \country{China}
}

\renewcommand{\shortauthors}{Weipeng Tan, et al.}

\begin{abstract}
Recent advances in Talking Head Generation (THG) have achieved impressive lip synchronization and visual quality through diffusion models; yet existing methods struggle to generate emotionally expressive portraits while preserving speaker identity. We identify three critical limitations in current emotional talking head generation: insufficient utilization of audio's inherent emotional cues, identity leakage in emotion representations, and isolated learning of emotion correlations. To address these challenges, we propose a novel framework dubbed as DICE-Talk, following the idea of disentangling identity with emotion, and then cooperating emotions with similar characteristics. First, we develop a disentangled emotion embedder that jointly models audio-visual emotional cues through cross-modal attention, representing emotions as identity-agnostic Gaussian distributions. Second, we introduce a correlation-enhanced emotion conditioning module with learnable emotion banks that explicitly capture inter-emotion relationships through vector quantization and attention-based feature aggregation. Third, we design an emotion discrimination objective that enforces affective consistency during the diffusion process through latent-space classification. Extensive experiments on MEAD and HDTF datasets demonstrate our method's superiority, outperforming state-of-the-art approaches in emotion accuracy while maintaining competitive lip-sync performance. Qualitative results and user studies further confirm our method's ability to generate identity-preserving portraits with rich, correlated emotional expressions that naturally adapt to unseen identities.
\end{abstract}




\onecolumn{%
    \renewcommand\onecolumn[1][]{#1}%
    \maketitle
    \begin{center}
	\centering
	\captionsetup{type=figure}
	\includegraphics[width=0.9\linewidth]{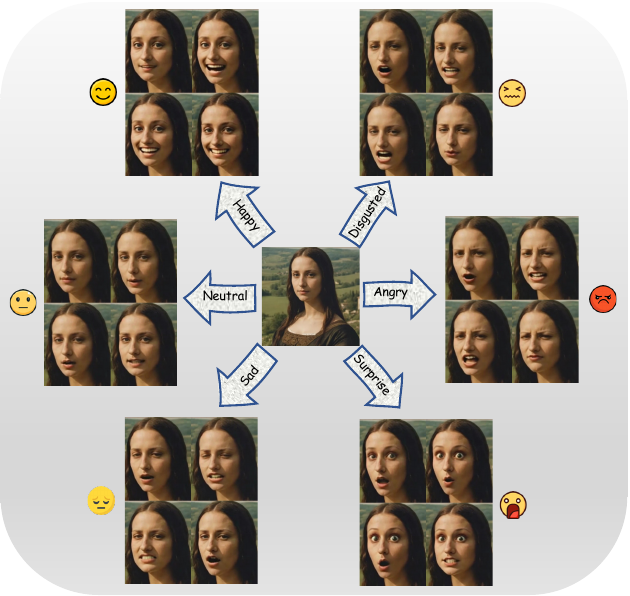} 
	\vspace{-3mm}
	\captionof{figure}{Our DICE-Talk, as an emotional THG method based on diffusion models, can generate various emotions while well preserving the identity characteristic, thus largely benefiting the real-life application of digital human.}
	\label{fig:intro}
    \end{center}%
}

\section{Introduction \label{sec:intro}}

Recent advancements in generative models have shed light on generating high-quality and realistic videos under various controlling conditions such as texts~\cite{rombach2022ldm}, images~\cite{blattmann2023svd,huang2024learning,pan2025earthsynth}, videos~\cite{wu2023tune}, etc. Among all the different subtasks of video generation, Talking Head Generation (THG), as a human-centric task which aims to generate videos of talking heads guided by conditions such as speech and images, has emerged as a significant problem due to its wide application in scenarios such as digital humans, film production, virtual reality. Despite importance, the extreme challenges result from its low tolerance to artifacts in general and its demand of high fidelity in lip shapes, facial expressions, and head motions. 

Following the commonly used generative models, GAN-based THG methods~\cite{prajwal2020wav2lip, zhang2023sadtalker} have achieved remarkable results in generating high-resolution videos through adversarial training between generators and discriminators, particularly in terms of visual quality and lip-sync accuracy. Diffusion model-based THG methods~\cite{shen2023difftalk,tian2024emo}, on the other hand, excel in generating high-quality and high-resolution images and videos, and outperform the GANs in terms of stability and consistency, thus becoming the mainstream methods for THG. These methods largely facilitate THG by strengthening the explicit controlling conditions such as facial keypoints and head motion sequences. Yet, a critical gap persists: \textit{none of these methods holistically model emotion as a controllable, multimodal signal}. Ignoring such a condition forces models to generate "emotionally flat" outputs, limiting their applicability in fields like mental health support, where emotional resonance is critical.

While it is straightforward to transfer the previous exploration on emotion control such as EAMM~\cite{ji2022eamm} and StyleTalk~\cite{ma2023styletalk} to diffusion backbones, these methods are generally restricted with three main omissions leading to incomprehensive control. (1) \textbf{Wrongly dominant visual emotion.} EAMM and StyleTalk propose adopting videos as the emotion source. However, audio as the elementary input of THG contain rich affective emotion cues such as the prosody and spectral tilt. For example, one can easily tell apart if someone is happy or sad by only listening to him speaking. Lack of such information can severely haunt the comprehensiveness of the emotion representation. (2) \textbf{Leaked identity characteristics.} Existing methods often fail to disentangle emotion from speaker identity, allowing unintended facial features (e.g., unique expressions or micro-movements) to leak into the generated output. This issue stems from treating emotions as fixed, universal attributes while overlooking the identity-specific volatility. For instance, the same emotion (e.g., happiness) may manifest differently across individuals due to unique facial structures or habitual mannerisms. Without explicit modeling of this volatility, transferred emotions may retain source-specific artifacts, compromising both generalization and fidelity when applied to novel identities. Failure to take into account such a fact degrades the generalizability of emotion transfer, especially when adapting to unseen identities. (3) \textbf{Isolated emotion learning.} Previous methods generally produce emotion representation based on single source video without extra knowledge. However, different emotions are actually correlated with each other. For example, learning emotions like sad and contempt can help model better understand other negative emotions such as disgusted. Such correlation, anyhow, is yet to be explored. 

To this end, we propose a novel framework named DICE-Talk, which can effectively extract well-disentangled emotion features with the assistance of audio information through a self-supervised method, and apply them to the generation of talking head videos in a manner suitable for diffusion models. This approach not only improves the overall quality of the generated videos, ensuring better synchronization and control but also accurately transfers facial expressions and individualized details to new faces.

Specifically, our DICE-Talk focuses on two main problems, i.e. extracting emotion embeddings and controlling diffusion models with such embeddings. For emotion embedding extraction, we propose the novel disentangled emotion embedder as an alternative to the previous methods such as StyleTalk based on the transformer backbone. Concretely, the audio and visual information of each video interacts with each other in the transformer emotion encoder, which models the emotion-related information of this video as a Gaussian distribution with predicted mean and standard deviation, aiming to simultaneously learn the general emotion representation and the identity-specific volatility of this video. Through contrastive learning, the extracted features exhibit significant clustering across different identities and emotions, not only helping the model better understand the video content but also providing an effective way to capture and express the emotion of individuals. 

In order to control the denoising process with the well learned emotion representation, we propose leveraging an emotion bank module to fully explore the intrinsic correlation between different emotions to boost each other. Specifically, this emotion bank first imports knowledge from the extracted emotion representations, from which the informative ones are then aggregated and incorporated in the diffusion UNet through cross attention. This enables the model to actively utilize the emotion source from the training set, which leads to strong generalization ability of the trained model. To encourage the model to be better controlled by the emotion condition, we further utilize an emotion discrimination objective apart from the noise prediction loss. Our objective incorporates an emotion discriminator trained to classify affective states from intermediate noisy latents of the diffusion UNet, ensuring precise control without sacrificing lip-sync accuracy or visual fidelity.

To validate the effectiveness of our proposed method, we conduct extensive experiments and comparisons on the MEAD~\cite{wang2020mead}, HDTF~\cite{zhang2021hdtf} and out-of-domain datasets. Our method significantly outperforms other competitors across multiple metrics, including FVD~\cite{unterthiner2018fvd}, FID~\cite{heusel2017fid}, SyncNet (Sync-C and Sync-D)~\cite{chung2017syncnet} and Emo-Scores. In addition to quantitative evaluation, we also perform comprehensive qualitative assessments. The results indicate that our method can generate highly natural and expressive talking videos and produce different emotions or even multiple changes in expressions within the same video according to user needs, achieving satisfactory visual effects.

Overall, our contributions are summarized as follows: 
(1) We propose DICE-Talk, an audio-driven diffusion-based THG framework that disentangles speaker identity from emotional expressions. This enables realistic talking head generation with rich emotions while preserving identity fidelity. 
(2) We introduce a cross-modal disentangled emotion embedder, representing emotions as identity-agnostic Gaussian distributions to address identity leakage and leverage audio's inherent prosodic cues. 
(3) A correlation-enhanced emotion conditioning mechanism is proposed, which leverages a learnable emotion bank to capture inter-emotion relationships.
\section{Related Work}
\label{sec:related_works}

\textbf{GAN-Based Talking Head Generation.}
There has been significant research on GAN-based methods for person-generic audio-driven talking head generation. Early methods~\cite{prajwal2020wav2lip,cheng2022videoretalking,wang2023lipread} achieved lip synchronization by establishing a discriminator that correlates audio with lip movements. Other approaches~\cite{zhou2020makeittalk,zhou2021pcavs,wang2021audio2head,wang2022one,zhang2023sadtalker} generated portrait videos by mapping audio to key facial information, such as landmarks, key points, or 3D Morphable Model (3DMM)~\cite{blanz1999morphable} coefficients, before rendering the final frame. However, due to the limitations of GANs in terms of generative capacity, the results produced by these methods often suffer from artifacts like pseudo-textures or restricted motion ranges.

\noindent\textbf{Diffusion Model-Based Talking Head Generation.}
Recently, there has been a surge of research ~\cite{shen2023difftalk,xie2024xportrait,wei2024aniportrait,yang2024megactor,cui2024hallo2, ji2024sonic, jiang2024loopy} utilizing diffusion models to achieve high-quality portrait videos. Among these, X-Portrait~\cite{xie2024xportrait} and MegActor~\cite{yang2024megactor} rely on the pose and expression from the source video to generate the target video, which limits their ability to produce videos based solely on audio. DiffTalk~\cite{shen2023difftalk} was the first to modify lip movements using audio and diffusion models, but it does not extend to driving other head parts. EMO~\cite{tian2024emo} was the first to leverage LDM~\cite{rombach2022ldm} and audio features to achieve overall motion in portraits. V-Express~\cite{wang2024vexpress} controls the overall motion amplitude by adjusting audio attention weights, while Hallo~\cite{xu2024hallo} designed a hierarchical module to regulate the motion amplitude of different regions. 
Sonic~\cite{ji2024sonic} focus on global audio perception to obtain better talking head results.
In summary, current audio-driven diffusion model approaches have not taken into account that each portrait should exhibit a corresponding emotion while speaking, which is essential for generating higher-quality portrait videos.

\noindent\textbf{Emotional Talking Head Generation.}
Previous research has explored several GAN-based methods ~\cite{ji2021evp,liang2022expressive} for extracting style information to apply in talking head generation. MEAD~\cite{wang2020mead} and Emotion~\cite{sinha2022emotion} directly inject style labels into the network to drive the corresponding emotions. GC-AVT~\cite{liang2022expressive} and EAMM~\cite{ji2022eamm} map the facial expressions in the source video to each frame in the target video. LSF~\cite{wu2021imitating} and StyleTalk~\cite{ma2023styletalk} employ 3D Morphable Models (3DMM) to extract facial information and construct style codes that drive the desired styles. 
EAT~\cite{gan2023eat} achieves emotion control through parameter-efficient adaptations.
Building on these approaches, our framework is first to disentangle identity and cooperate emotion to generate better emotion features into diffusion model and produce high-quality emotional talking portrait videos. 
\section{Method}
\begin{figure*}[t!]
	\centering
	\includegraphics[width=1.7\columnwidth]{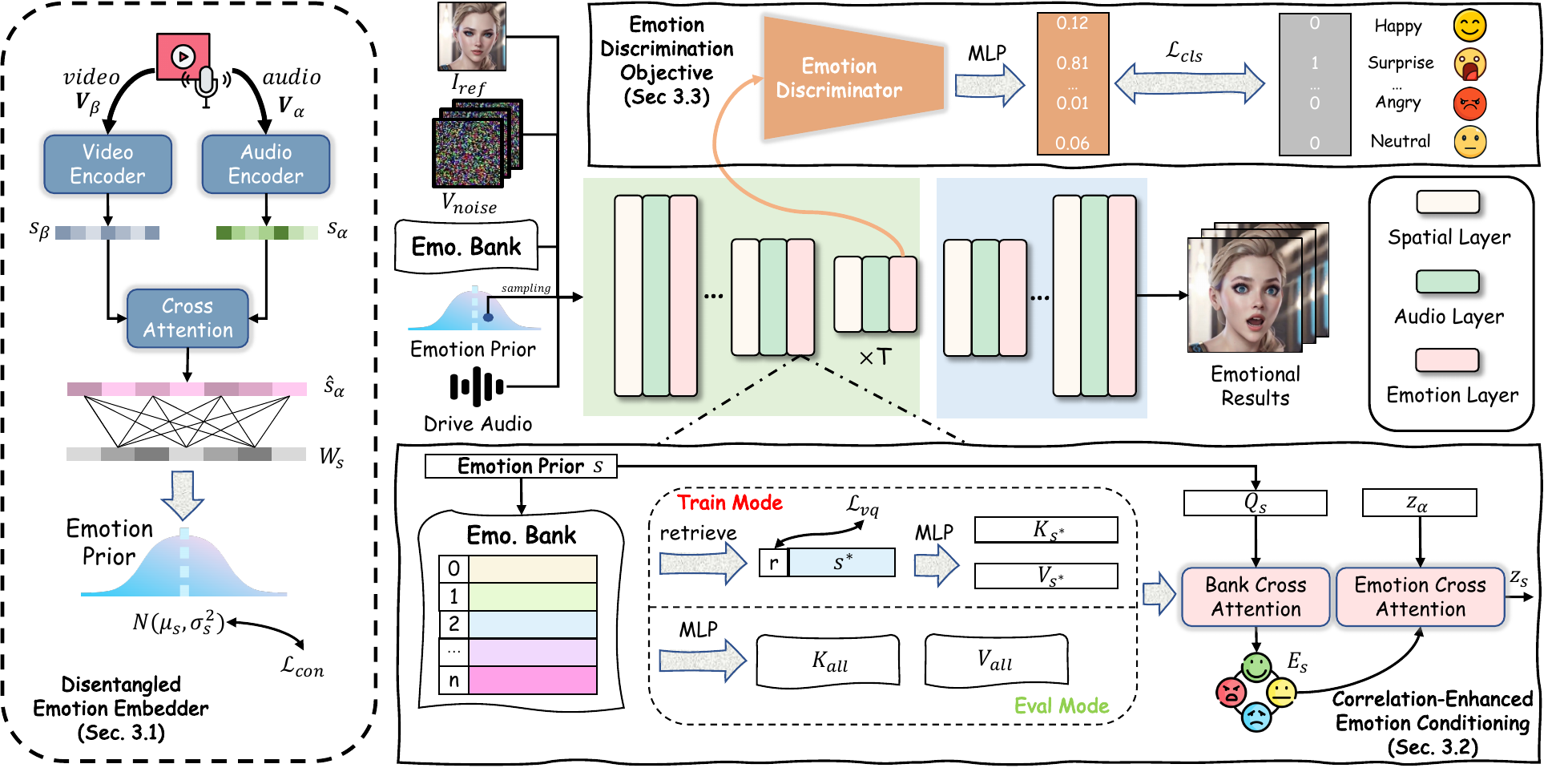} 
	\vspace{-3mm}
	\caption{\textbf{Framework of DICE-Talk. Our method comprises three key components: disentangled emotion embedder, correlation-enhanced emotion conditioning, and emotion discrimination objective. These architectural elements work synergistically to decouple identity representations from emotional cues while preserving facial articulation details, thereby generating lifelike animated portraits with emotionally nuanced expressions.}
	} 
	\label{fig:framework}
 \vspace{-4mm}
\end{figure*}
\textbf{Problem Formulation.}~
The goal of THG is to generate a talking head video under the control of a reference portrait image, audio, and emotion prior. Among these conditions, the reference portrait image provides the background and facial identity, the audio guides the overall head motion and lip movements. In addition, we propose a disentangled emotion embedder to extract the emotion prior from the visual and audio content of an emotion reference video, which is used to determine facial emotion.

\noindent\textbf{Preliminaries.}~
In DICE-Talk, we employ a Stable Video Diffusion (SVD)~\cite{blattmann2023svd} with EDM~\cite{karras2022edm} framework to generate video frames. The SVD uses a diffusion and denoising process in the latent space via a Variational Autoencoder (VAE). Given a video $x \in \mathbb{R}^{N\times 3 \times H\times W}$, it maps each frame into the latent space, encoding the video as $z=E(x)$, which helps maintain visual quality while reducing computational cost. During the diffusion process, Gaussian noise $\epsilon \sim \mathcal{N}(\mathbf{0},\mathbf{I})$ is gradually introduced into the latent $z$, degrading it into complete noise $z_T\sim\mathcal{N}(\mathbf{0},\mathbf{I})$ after $T$ steps. In the reverse denoising process, the target latent $z$ is iteratively denoised from the sampled Gaussian noise using the diffusion model and then decoded by the VAE decoder $D$ into the output video $x=D(z)$. During training, given the latent $z_0 =E(x_0 )$ and condition $c$, the denoising loss is:
\begin{equation}
\mathcal{L}_{{denoising}}=\mathbb{E}_{\mathbf{z}_t,\boldsymbol{\epsilon},\mathbf{c},t}\|\boldsymbol{\epsilon}_\theta(\mathbf{z}_t,\mathbf{c},t)-\boldsymbol{\epsilon}_t\|^2.
\label{eq:denoise}
\end{equation}
Among them, $z_t$ represents the noisy latent variables at timestep $t\in[1,T]$, and $\epsilon_t$  is the added noise. $\epsilon_\theta$ is the noise predicted by the UNet model, modified using an attention mechanism with parameters $\theta$. This model employs a cross-attention mechanism to fuse the condition $c$ with the latent features $z_t$, thereby guiding the video generation.

\noindent\textbf{Overview.}~
As depicted in Figure~\ref{fig:framework}, our DICE-Talk consists of three important designs: (1) A disentangled emotion embedder that extracts emotions from video and audio inputs, representing them as a Gaussian distribution. (2) A correlation-enhanced emotion conditioning module that stores emotion features in a bank and injects them into a diffusion model. (3) An emotion discrimination objective that improves emotion expression by classifying emotions during denoising. The diffusion process combines a reference image, audio features, emotion data, and the emotion bank to guide facial animation. Audio features come from Whisper-Tiny~\cite{radford2023whisper}, and following~\cite{ji2024sonic}, we use an attention layer for processing. Finally, after denoising, a VAE decoder generates expressive portrait frames with the target emotion.

\subsection{Disentangled Emotion Embedder}\label{Disentangled_emo_emb}

In order to make full use of the related factors indicating emotion of a speaker to achieve the well-disentangled emotion embeddings, we adopt a novel framework named disentangled emotion embedder. Concretely, for a $N$-frame video clip $\mathbf{V}$ consisting of the frame set $\mathbf{V}_{\beta}=\{\mathbf{I_i}\}_{i=1}^N$ and its corresponding audio $\mathbf{V}_{\alpha}$, we first extract the audio and visual features respectively. For the visual information, we follow StyleTalk to encode $\mathbf{V}_{\beta}$ into feature sequence $s_\beta \in \mathbb{R}^{N\times d_s}$. Beyond that, we further stack pretrained Whisper-Tiny~\cite{radford2023whisper} with additional learnable transformer layers to extract sequential audio feature $s_\alpha \in\mathbb{R}^{N\times d_s}$ from $\mathbf{V}_{\alpha}$. 

After achieving features for each modality, our disentangled emotion embedder further merges audio information into the visual one through cross-attention mechanism. Specifically, for each layer, $s_\alpha$ is updated as follows:
\begin{align}
    Q_\alpha &= W_Q(s_\alpha); K_\beta = W_K(s_\beta); V_\beta = W_V(s_\beta), \\
    \hat{s}_\alpha &= s_\alpha + \lambda \mathrm{CA}(Q_\alpha, K_\beta, V_\beta),
\end{align}
where $\mathrm{CA}(\cdot, \cdot, \cdot)$ denotes cross attention. In this way, $s_\alpha$ is entitled with the awareness of dual modalities related to human emotion.

With the fused $\hat{s}_\alpha$, we then propose modeling the emotion condition for each video as a Gaussian distribution. Specifically, an attention-based aggregation strategy is employed on $\hat{s}_\alpha$ as follows:
\begin{align}
    \mu_s &= \mathrm{softmax}(W_s\hat{s}_\alpha)\cdot \hat{s}_\alpha^T,\\
    \sigma_s^2 &= \mathrm{softmax}(W_s \hat{s}_\alpha)\cdot(\hat{s}_\alpha^T-\mu_s)^2,\\
    s &= \mu_s + \sigma_s \cdot \epsilon, \quad \epsilon \sim \mathcal{N}(\mathbf{0}, \mathbf{I}), \label{eq:emotion}
\end{align}
where $W_s\in\mathbb{R}^{1\times d_s}$ is a trainable parameter, $\mu_s, \sigma_s^2$ denotes the mean and variance of the learned emotion prior $s$.

Compared with StyleTalk, our proposed model mainly enjoys the following merits: (1) \textbf{Unified audio-visual emotion modeling.} As mentioned in Sec.~\ref{sec:intro}, audio should be taken as an important indicator of emotion. We design a specific structure to handle these complex data, considering both visual and audio information, leading to stronger emotion embedding. (2) \textbf{Identity-emotion disentanglement.} Models such as StyleTalk cannot learn to fully disentangle identity information from emotion features due to insufficient capability via solely modeling deterministic emotion features. Therefore, such identity-related features would be leaked into the emotion embedding. Our method, on the other hand, model the emotion prior as a Gaussian distribution that is more representative, thus learning a better disentangled sequential embeddings.

\noindent\textbf{Embedder Training.} Essentially, codes with similar emotions should cluster together in the emotion space. Therefore, we apply contrastive learning to the emotion priors by constructing positive pairs $(s, s^p)$ with the same identity and emotion, and negative pairs $(s, s^n)$ with different identities or emotions. Then, the InfoNCE loss~\cite{chen2020simclr} with similarity metric $\zeta$ is enhanced between positive and negative sample pairs:
\begin{align}
    \omega(\tilde{s}) &= \exp(\zeta(s, \tilde{s}) / \tau), \\
    \mathcal{L}_{con} &= -\log\left(\frac{\omega(s^p)}{\omega(s^p)+\sum_{s^n\in\mathcal{S}^n}\omega(s^n)}\right), \label{eq:infonce}
\end{align}
where $\tau$ denotes a temperature parameter, $\mathcal{S}^n$ denotes all negative samples for $s$, and $\zeta(s_i, s_j)$ is the cosine similarity between sample pairs $(s_i, s_j)$.

In the training of the disentangled emotion embedder, we directly train all parameters of this lightweight model. Meanwhile, we use a random dropout trick when inputting emotion features $s_{\beta}$ and audio features $s_\alpha$ by setting some of the emotion features $s_\beta$ or audio features $s_\alpha$ to zero. This allows the model to obtain the emotion prior through a single modality.

\subsection{Correlation-Enhanced Emotion Conditioning}\label{stylebank}
In order to control the emotion of talking head videos, a straightforward way is to directly inject $s$ achieved in Eq.~\ref{eq:emotion} into SVD through cross attention, following other conditional diffusion models such as Sonic~\cite{ji2024sonic}. However, while $s$ can well depict the emotion condition of each video clip, it still cannot well handle the internal correlation among different emotions, which leads to limited performance.

To address the problem, we propose a correlation-enhanced emotion conditioning process via an emotion bank attention layer. Rather than directly leveraging information contained in $s$ as condition, we instead utilize an emotion bank $\mathcal{C}\in \mathbb{R}^{K\times d_s}$ to first refine the emotion information. Specifically, during training, following previous works on discrete encoding, a emotion memory $s^*$ is retrieved from $\mathcal{C}$ based on the extracted emotion embedding $s$, whose information is further aggregated according to $s$:
\begin{align}
    s^* &= \mathcal{C}_{k};\ k=\mathrm{arg}\min_{i}\|s-\mathcal{C}_i\|_2^2, \\
    Q_s &= W_Q(s); K_{s^*} = W_K(s^*); V_{s^*} = W_V(s^*), \label{eq:bank-ca-proj}\\
    E_S &= \mathrm{CA}(Q_s, K_{s^*}, V_{s^*}), \label{eq:bank-ca}
\end{align}
By training $\mathcal{C}$ with objectives such as commitment loss~\cite{van2017vqvae}, it learns to memorize the shared emotion-related knowledge among the training set. Intuitively, embeddings with the same emotion would be clustered into similar $s^*$, which in turn reduces the effect of identity leakage. Consequently, the retrieval process matches the appropriate feature distribution for the specific input, thereby expressing the corresponding emotion. Additionally, this method can express entirely new emotions through combinations of different emotional features, providing better generalization. Moreover, the training process can encourage $W_Q,W_K,W_V$ to discover latent subspaces where emotion correlations are salient, which further benefits the generation process: After achieving a well-trained $\mathcal{C}$, the retrieval process can be skipped during inference, i.e. replacing $s^*$ in Eq.~\ref{eq:bank-ca-proj} with $\mathcal{C}$, i.e.
\begin{align}
    Q_s &= W_Q(s); K_{all} = W_K(\mathcal{C}); V_{all} = W_V(\mathcal{C}), \label{eq:bank-ca-proj-infer}\\
    E_S &= \mathrm{CA}(Q_s, K_{all}, V_{all}). \label{eq:bank-ca-infer}
\end{align}
In this way, the cross attention as in Eq.~\ref{eq:bank-ca} allowing $s$ to attend to multiple relevant prototypes (e.g., both "anger" and "disgust" for aggressive expressions).

\noindent\textbf{Emotion Projection.}~
To utilize the emotion feature $E_s$ obtained in Sec.~\ref{stylebank} and guide the denoising process, then $E_s$ is injected into the diffusion UNet through an additional emotion attention layer, where it interacts with other features via a cross-attention mechanism to supplement additional facial details such as expressions.
The information of the emotion feature $E_s$  can be injected into the spatial and temporal cross-attention layers to provide spatial knowledge as follows:
\begin{equation}
z_{s}= z_a + \mathrm{CA}\left(Q(z_a),K(E_s),V(E_s)\right),
\label{eq:style}
\end{equation}
where $z_a$ is the spatial latent features after being injected with reference attention and audio attention, and $z_s$ is the adjusted spatial features guided by emotion prior spatial-aware level.

\subsection{Compositional Diffusion Adaptation}\label{diff_adapt}

To finetune the newly introduced parameters in SVD, we adopt a compositional objective as follows
\begin{equation}
    \mathcal{L} = \mathcal{L}_{denoising} + \lambda\mathcal{L}_{cls} + \mathcal{L}_{vq}, 
\end{equation}
where $\mathcal{L}_{denoising}$ denotes the noise prediction loss as in Eq.~\ref{eq:denoise}, $\mathcal{L}_{cls}$ and $\mathcal{L}_{vq}$ respectively denote emotion discrimination objective and emotion bank learning objective, which will be detailed below.

\noindent\textbf{Emotion Discrimination Objective.} To promote the model to generate more precise emotions, we introduce the emotion discrimination objective. Formally, denoting $f_t \in \mathbb{R}^{(C\times H \times W)\times d}$ as the intermediate features extracted from the U-Net at timestep $t$, where $C, H, W$ represent channel, height, and width dimensions respectively. These features are first compressed by an adaptive pooling layer $\phi$ into a fixed-dimensional vector $v_t=\phi(f_t)$, which is then projected through an MLP classifier $\psi$ to obtain the emotion probability distribution $p_t=\mathrm{softmax}\left(\psi(v_t)\right)$ The auxiliary cross-entropy loss $\mathcal{L}_{cls}$ is computed as:
\begin{equation}
\mathcal{L}_{cls}=-\frac{1}{T}\sum_{t=1}^T\sum_{c=1}^Ny_c\log p_{t,c}
\end{equation}
where $T$ is the total number of sampled timesteps, $N$ denotes the number of emotion categories, $y_c$ is the ground-truth one-hot label for category $c$ , and $p_{t,c}$ represents the predicted probability for category $c$ at timestep $t$. $\mathcal{L}_{cls}$ guides the model to focus on the accuracy of emotions during content generation, thus helping the model better capture emotion-related information in the noisy latent space.

\noindent\textbf{Emotion Bank Learning.}~
The emotion bank $\mathcal{C}$ is optimized through a vector quantization loss $\mathcal{L}_{vq}$, which jointly refines both the emotion prototypes and their association with input features. Given an input emotion prior and the codebook $C$ , the loss function operates bidirectionally: The codebook commitment loss $\|\mathrm{sg}[s] - C_k\|_2^2$ updates only the selected prototype $C_k$ (where $k=\mathrm{arg}\min_{i}\|s-C_i\|_2^2$) toward the input feature using stop-gradient operation $\mathrm{sg}[\cdot]$, while the feature alignment loss $\|s - \mathrm{sg}[C_k]\|_2$ drives the input features toward their nearest prototype without modifying the codebook. The composite loss:
\begin{equation}
\mathcal{L}_{vq} = \|\mathrm{sg}[\mathbf{s}] - \mathbf{C}_k\|_2^2 + \beta\cdot\|\mathbf{s} - \mathrm{sg}[\mathbf{C}_k]\|_2^2,
\end{equation}
where $\beta $ is the commitment weight.

\section{Experiments}
\begin{figure*}[t!]
	\centering
	\includegraphics[width=1.75\columnwidth]{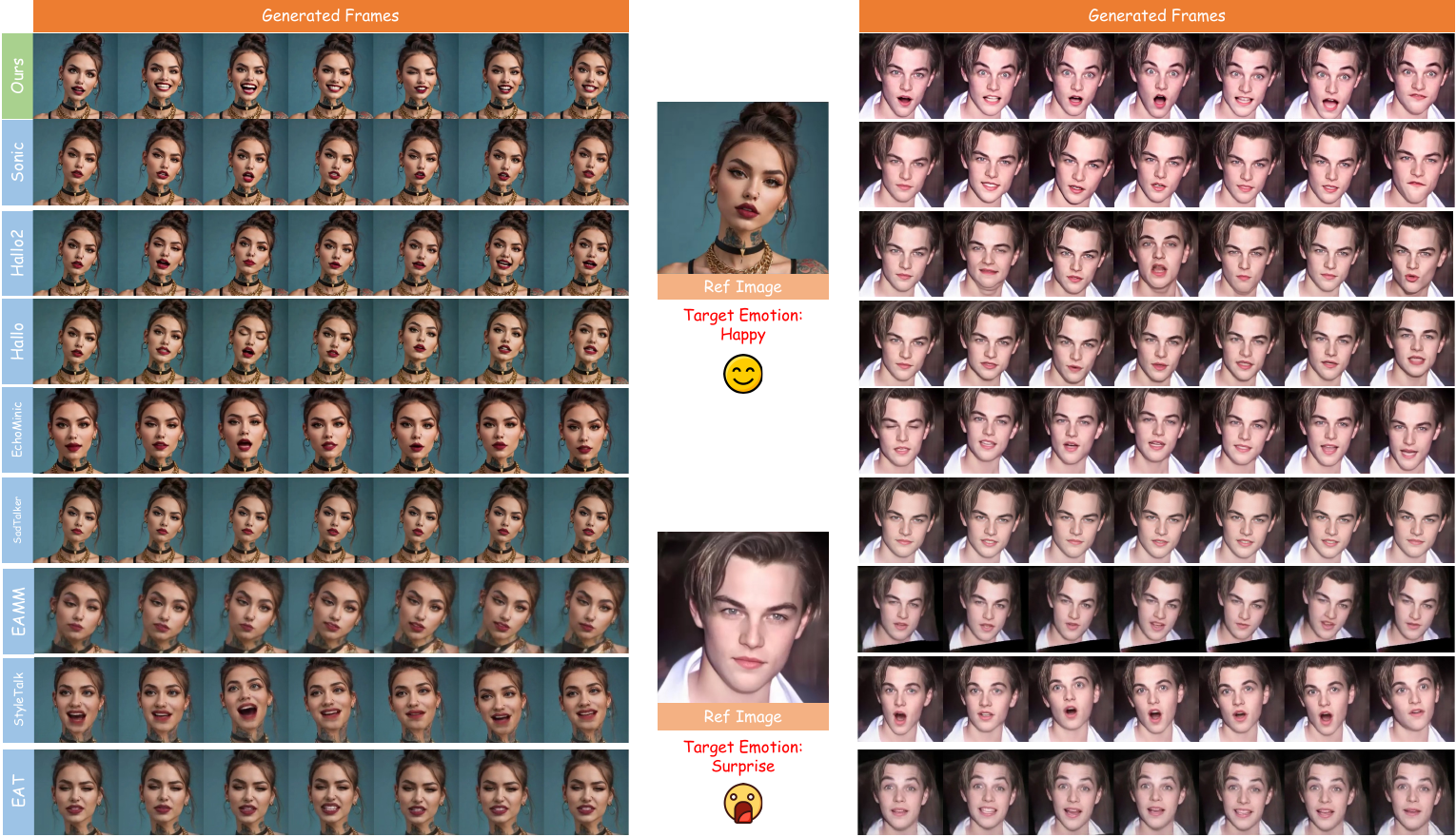} 
	\vspace{-3mm}
	\caption{\textbf{Visual comparisons with recent state-of-the-art taking head generation methods.} Our method can obtain more accurate emotional expressions. Besides, we propose the full video comparison in supplementary materials to represent the capability of our method on sync, naturalness and stability.}
	\label{fig:compare}
 \vspace{-4mm}
\end{figure*}

\begin{figure*}[t!]
	\centering
	\includegraphics[width=1.75\columnwidth]{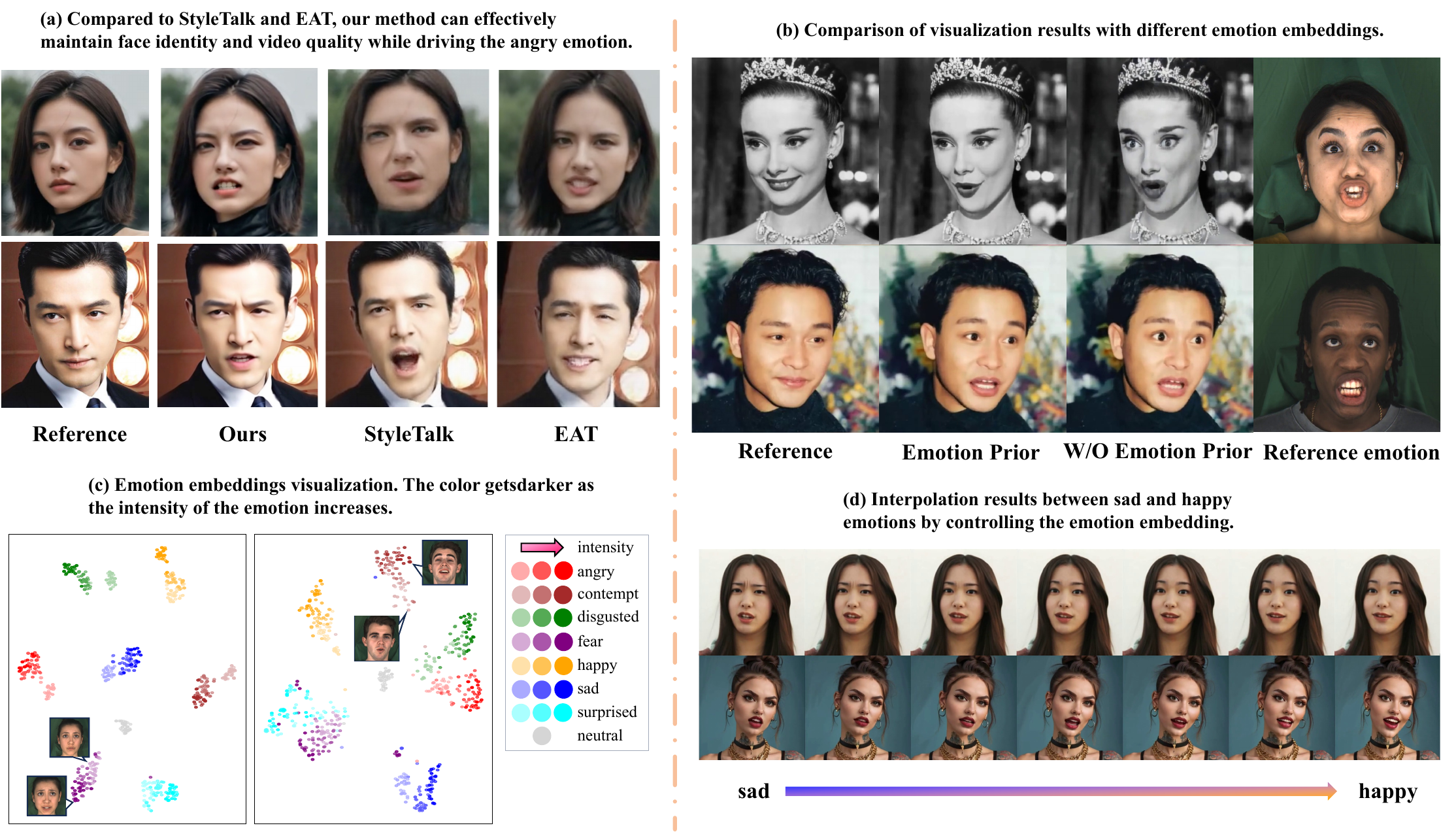} 
	\vspace{-3mm}
	\caption{\textbf{Visual results of user and ablation studies.} (a): Visual comparisons with StyleTalk and EAT. (b): Comparison of visualization results with different emotion embeddings. (c): Emotion embeddings visualization. (d): Interpolation results between sad and happy emotions by controlling the emotion embedding.
}
	\label{fig:ablation_all}
 \vspace{-2mm}
\end{figure*}

\subsection{Experiments Setting}

DICE-Talk~is implemented using PyTorch~\cite{paszke2019pytorch} and optimized with Adam~\cite{kingma2014adam}.
The disentangled emotion embedder is trained on the MEAD~\cite{wang2020mead} and HDTF~\cite{zhang2021hdtf} datasets. 
During training, we consider samples with the same identity and emotion in MEAD as positive samples, and segments from the same video in HDTF as positive samples. Additionally, we will randomly dropout expression coefficients or audios, but they will not be zeroed out simultaneously.

The training dataset for the diffusion model includes both emotion dataset and regular video dataset, which are incorporated into the training process at a 1:1 ratio. We use MEAD dataset as the emotion dataset, while the regular video datasets include HDTF, VoxCeleb2~\cite{Chung_2018}, and others from the Internet.

The training of the diffusion model consists of only one stage, with the initialization of the spatial module and the temporal modules from stable-video-diffusion-xt-1-1~\cite{blattmann2023stable}. To separately enable different conditions during training, we manipulate the data such that 5\% of it drops audio, 5\% drops image, 5\% drops emotion prior, and 5\% drops all conditions.

For training and testing set splitting, we select 5 identities out of 46 for testing on MEAD. For HDTF, we randomly select 20 videos for testing. Precautions are taken to ensure that there is no overlap of character identities between the training and testing sets. To further evaluate generalization performance, we collected an additional out-of-domain dataset comprising 10 reference images and corresponding audio samples from diverse online sources for testing. 
We selected reference images of speakers with neutral emotions from the MEAD dataset and used videos of the same person with different emotions from the dataset to generate corresponding emotional videos. For the out-of-domain dataset, we randomly selected reference videos for different emotions. For the HDTF dataset, we used the original videos as emotional references to generate videos for the reference images. To ensure a fair comparison with EAT, we adopted a similar approach by preparing a fixed feature for each emotion. Specifically, we used the mean of the emotion priors from a large number of videos with the same emotion as the prompt feature for the corresponding emotion.
During inference, we utilize ~\cite{karras2022edm} as diffusion sampler with the denoising steps set as 25 for all diffusion-based methods.

\begin{table*}[t!]
\footnotesize
\centering
\caption{\textbf{Quantitative comparisons with the state-of-the-arts on the MEAD and Out of Domain test set.} The best results are \textbf{bold}, and the second are \underline{underlined}. Ours-V represents the emotion prior coming from a separate emotional reference video, while Ours-P represents the emotion prior as a fixed prompt feature we set for the corresponding expression.}
\label{tab:compare_mead_ood}
\vspace{-3mm}
\setlength{\tabcolsep}{0.9mm}{
\begin{tabular}{@{}ccccrrcccrrccc@{}}
\toprule
 &
 &
  \multicolumn{7}{c}{MEAD} &
  \multicolumn{5}{c}{Out of Domain Emotion Testset} \\ \cmidrule(l){3-9} \cmidrule(l){10-14} 
  Method &
  Condition &
  FVD$\downarrow$ &
  FID$\downarrow$ &
  \multicolumn{1}{c}{Sync-C$\uparrow$} &
  \multicolumn{1}{c}{Sync-D$\downarrow$} &
  Emo-Score$\uparrow$ &
  Neutral$\uparrow$ &
  Non-neutral$\uparrow$ &
  \multicolumn{1}{c}{Sync-C$\uparrow$} &
  \multicolumn{1}{c}{Sync-D$\downarrow$} &
  Emo-Score$\uparrow$ &
  Neutral$\uparrow$ &
  Non-neutral$\uparrow$ \\ \midrule
EAMM &
  audio + video &
  944.3041 &
  122.2852 &
  4.1973 &
  9.5079 &
  0.2220 &
  0.4007 &
  0.1624 &
  3.4361 &
  10.1884 &
  0.2411 &
  0.6106 &
  0.1179 \\
StyleTalk &
  audio + video &
  498.3980 &
  104.0158 &
  4.9014 &
  10.0728 &
  0.4323 &
  0.5119 &
  0.4058 &
  3.2781 &
  11.1976 &
  0.4722 &
  0.4200 &
  0.4895 \\
Ours-V &
  audio + video &
  382.1692 &
  54.0682 &
  6.2182 &
  8.7805 &
  0.4615 &
  0.4659 &
  0.4601 &
  \underline{6.2720} &
  \underline{8.0744} &
  \underline{0.5424} &
  0.6393 &
  \underline{0.5101} \\ \midrule
EAT &
  audio + prompt &
  629.4533 &
  117.4011 &
  \textbf{6.9590} &
  \textbf{8.2743} &
  \underline{0.4673} &
  0.3210 &
  \textbf{0.5160} &
  4.4362 &
  10.0130 &
  0.3865 &
  0.5323 &
  0.3379 \\
Ours-P &
  audio + prompt &
  377.0500 &
  \underline{52.9348} &
  6.1158 &
  8.9969 &
  \textbf{0.4865} &
  0.5373 &
  \underline{0.4696} &
  6.1751 &
  8.3802 &
  \textbf{0.5527} &
  0.6281 &
  \textbf{0.5276} \\ \midrule
SadTalker &
  audio &
  697.1468 &
  83.4242 &
  5.1855 &
  8.8352 &
  0.2198 &
  \textbf{0.7447} &
  0.0448 &
  5.0071 &
  8.6663 &
  0.2430 &
  \underline{0.7225} &
  0.0831 \\
V-Express &
  audio &
  537.5223 &
  61.8953 &
  3.1232 &
  10.6003 &
  0.2136 &
  0.5399 &
  0.1048 &
  2.9307 &
  10.8661 &
  0.2356 &
  0.5768 &
  0.1219 \\
AniPortait &
  audio &
  633.7429 &
  60.4030 &
  1.1205 &
  12.5940 &
  0.2179 &
  0.4755 &
  0.1320 &
  1.1171 &
  12.2194 &
  0.2356 &
  0.5486 &
  0.1313 \\
Hallo &
  audio &
  \textbf{326.7183} &
  55.9922 &
  6.0092 &
  8.6960 &
  0.2131 &
  0.5257 &
  0.1089 &
  6.1672 &
  8.1545 &
  0.2367 &
  0.6304 &
  0.1055 \\
Hallo2 &
  audio &
  478.0882 &
  57.6818 &
  6.3825 &
  8.6008 &
  0.2130 &
  0.5133 &
  0.1130 &
  6.1309 &
  8.2896 &
  0.2394 &
  0.5748 &
  0.1276 \\
EchoMinic &
  audio &
  616.0920 &
  74.9136 &
  5.7066 &
  9.0599 &
  0.2219 &
  \underline{0.5417} &
  0.1153 &
  4.8856 &
  9.2830 &
  0.2466 &
  \textbf{0.7237} &
  0.0875 \\
Sonic &
  audio &
  \underline{370.7403} &
  \textbf{52.2722} &
  \underline{6.7164} &
  \underline{8.4195} &
  0.2586 &
  0.4832 &
  0.1837 &
  \textbf{6.9257} &
  \textbf{7.3133} &
  0.2444 &
  0.6014 &
  0.1255 \\ \bottomrule
\end{tabular}}
\vspace{-3mm}
\end{table*}
\subsection{Quantitative Comparison}

\begin{table}[t!]
\centering
\caption{Quantitative comparisons with the state-of-the-arts on the HDTF test set.}
\label{tab:compare_hdtf}
\vspace{-4mm}
\resizebox{0.45\textwidth}{!}{
\begin{tabular}{@{}cccccc@{}}
\toprule
Method                    &  Condition                   & FVD$\downarrow$               & FID$\downarrow$             & Sync-C$\uparrow$          & Sync-D$\downarrow$          \\ \midrule
SadTalker               & audio                  & 450.5424          & 15.6347         & 5.7955          & 8.5125          \\
V-Express               & audio                  & 429.8870          & 11.2261         & 1.3750          & 12.5764         \\
AniPortait              & audio                  & 246.2741          & \underline{9.3376}    & 0.5623          & 13.5954         \\
Hallo                   & audio                  & \underline{188.8993}    & 9.9675          & 7.0955          & 7.7918          \\
Hallo2                  & audio                  & 261.6602          & 9.5827          & \underline{7.4964}    & 7.9497          \\
EchoMinic               & audio                  & 410.4968          & 38.1664         & 6.0018          & 9.2090          \\
Sonic                   & audio                  & \textbf{157.1348} & \textbf{9.0076} & \textbf{7.7676} & \textbf{7.3520} \\ \midrule
EAMM                    & audio + video          & 887.3187          & 68.4042         & 3.9795          & 9.6646          \\
StyleTalk               & audio + video          & 491.8348          & 52.7492         & 5.4025          & 9.3879          \\
Ours-V                  & audio + video          & 258.3758          & 10.3833         & 6.8049          & 8.3610          \\ \midrule
EAT                     & audio + prompt         & 586.7120          & 56.3571         & 7.3534          & \underline{7.6736}    \\
Ours-P                  & audio + prompt         & 223.9416          & 10.0854         & 7.3218          & 7.8157          \\ \bottomrule
\end{tabular}}
\vspace{-5mm}
\end{table}
We compare with several previous works, including SadTalker~\cite{zhang2023sadtalker}, AniPortrait~\cite{wei2024aniportrait}, V-Express~\cite{wang2024vexpress}, Hallo~\cite{xu2024hallo}, Hallo2~\cite{cui2024hallo2}, EchoMinic~\cite{chen2024echomimic}, Sonic~\cite{ji2024sonic}, EAMM~\cite{ji2022eamm}, StyleTalk~\cite{ma2023styletalk}, and EAT~\cite{gan2023eat}. Among these, EAMM, StyleTalk, and EAT are also focused on emotion modeling.

To demonstrate the superiority of the proposed method, we evaluate the model using several quantitative metrics. We utilize the Fréchet Inception Distance (FID)~\cite{heusel2017fid} to assess the quality of the generated frames, and further employ the Fréchet Video Distance (FVD)~\cite{unterthiner2018fvd} for video-level evaluation. To evaluate lip-sync accuracy, we use the Sync-C and Sync-D of SyncNet~\cite{chung2017syncnet}. 
To assess the accuracy of the generated facial expressions, we use the Emo-Score~\cite{RYUMINA2022emoscore}. Specifically, we selected four easily recognizable emotions—neutral, happy, surprised, and angry (emotions like sad and fear are inherently more easily confused with other emotions)—to evaluate the effectiveness of emotion generation. The average of these scores will be used as the overall Emo-Score. Additionally, we will compare the average scores of neutral and non-neutral emotions.

As shown in Table~\ref{tab:compare_mead_ood}, our method ranks among the top in metrics on both the MEAD and out-of-domain datasets. Our method achieved the best emotion scores, reflecting that our approach generates vivid and accurate facial expressions through emotion priors. In the quality assessment of videos, single-frame images, and lip-sync accuracy, our method is close to the current best methods, as evidenced by the FVD, FID, and SyncNet scores. The results in Table~\ref{tab:compare_hdtf} further indicate that our method maintains high visual quality even on datasets with no obvious emotions like HDTF, which do not have obvious emotions. These results demonstrate that our method can ensure video generation quality while achieving diversity and accuracy in emotional expression.

\subsection{Qualitative Comparison}
\begin{figure*}[t!]
	\centering
	\includegraphics[width=1.8\columnwidth]{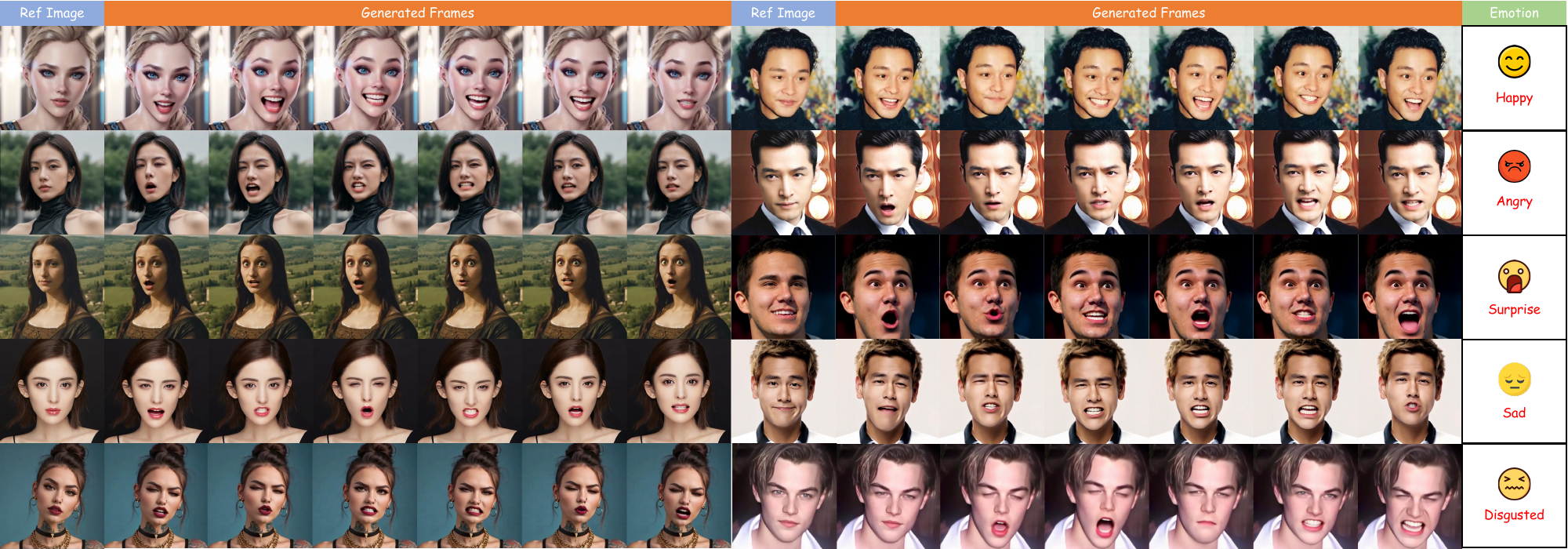} 
	\vspace{-2mm}
	\caption{\textbf{Our visual results with different emotions and identities on out of domain dataset.}
	} 
	\label{fig:emos}
 \vspace{-2mm}
\end{figure*}

In Figure~\ref{fig:compare}, we present a visual comparison of our method with other methods.
Our method successfully generates videos with rich expressions and natural movements on open datasets, demonstrating strong robustness. Analysis results show that our method exhibits significant advantages in emotional expressiveness and visual quality. Compared to recent THG methods, our model achieves diverse emotional expressions while maintaining facial details, whereas ordinary models often appear monotonous and lack subtlety in emotional expression. Compared to emotional models, our method has advantages in the accuracy and naturalness of emotional expression, enabling smooth transitions between emotions, while other models may appear stiff during emotional transitions. Additionally, our model supports generating higher-resolution images, resulting in significantly better clarity and visual quality.

\begin{table}[t!]
\centering
\caption{User study comparison on out of domain dataset.}
\label{tab:user-std}
\vspace{-3mm}
\begin{tabular}{@{}ccccc@{}}
\toprule
Method    & Emotion         & Lip Sync        & Quality         & Smoothness      \\ \midrule
EAMM      & 1.3482          & 1.4286          & 1.3036          & 1.4643          \\
StyleTalk & 2.2679          & 1.7857          & 1.5893          & 1.7321          \\
EAT       & 2.2321          & 2.0357          & 1.6071          & 1.8571          \\
Sonic     & 2.0536          & \textbf{3.9911} & \textbf{3.9911} & 3.9732          \\ \midrule
Ours      & \textbf{3.7768} & 3.9286          & 3.9375          & \textbf{4.0357} \\ \bottomrule
\end{tabular}
\vspace{-4mm}
\end{table}

\subsection{User Study}
We conducted a subjective evaluation of the results on open datasets, assessing the comparative methods along four dimensions: emotional expression, lip-sync accuracy, video quality, and video smoothness. 20 participants rated the results of five comparative methods on a scale of 1 to 5 in a total of 40 video sets. As shown in Table~\ref{tab:user-std}, our DICE-Talk significantly outperformed others in the dimensions of emotional expression and video smoothness, with a notable 66.5\% improvement in emotional expression. Although StyleTalk and EAT achieved good results in objective metrics, they compromised the original identity and reduced visual quality while generating emotions, leading to lower subjective emotion scores. Figure~\ref{fig:ablation_all}(a) illustrates two clear examples. In terms of lip-sync accuracy and video quality, our method is very close to the state-of-the-art Sonic and far surpasses other methods that model emotions. 
Video results from the user study are included in the supplementary materials.

\begin{table}[t!]
\centering
\caption{Ablation studies on our method.}
\label{tab:ablation}
\vspace{-4mm}
\resizebox{0.44\textwidth}{!}{
\begin{tabular}{@{}ccccc@{}}
\toprule
Methods                     & F-SIM$\uparrow$           & Emo-Score$\uparrow$       & Neutral$\uparrow$          & Non-neutral$\uparrow$      \\ \midrule
w/o emotion bank            & \textbf{0.8775} & 0.4781          & \textbf{0.6986} & 0.4047          \\
w/o $\mathcal{L}_{cls}$                & 0.8502          & 0.4762          & 0.6636          & 0.4138          \\
learnable CA             & 0.8622          & 0.5293          & 0.6754          & 0.4806          \\
w/o emotion prior & 0.8523          & 0.4992          & 0.6539          & 0.4477          \\ \midrule
Ours                        & 0.8704          & \textbf{0.5424} & 0.6393          & \textbf{0.5101} \\ \bottomrule
\end{tabular}}
\vspace{-3mm}
\end{table}

\subsection{Ablation Study}

\begin{table}[t!]
\small
\centering
\caption{Comparison of clustering strength on different emotions. The higher value means the better clustering effects.}
\label{tab:cluster_emo}
\vspace{-3mm}
\resizebox{0.38\textwidth}{!}{%
\setlength{\tabcolsep}{0.6mm}{
\begin{tabular}{@{}cccccc@{}}
\toprule
Input         & angry & contempt & disgusted & happy \\ \midrule
audio         & 2.44  & 2.12     & 2.51 & 2.49  \\
video         & 6.25  & 4.89     & 7.45 & 8.07  \\
video + audio & \textbf{6.57}  & \textbf{5.64}     & \textbf{7.63} & \textbf{8.38}  \\ \bottomrule
\end{tabular}}}
\vspace{-6mm}
\end{table}

\noindent\textbf{Emotion Conditioning.}~
We analyzed the impact of different emotion control methods on emotion conditioning effectiveness. We compared three approaches: without using emotion bank, without using the emotion discrimination objective, and compute cross attention between learnable embeddings and the emotion prior. For a more detailed evaluation, we additionally included the F-SIM~\cite{tian2024emo} to assess the facial similarity between the generated video and the reference image. The experiments show that our method effectively enhances emotional expression capabilities while better preserving the identity of the reference image. Table~\ref{tab:ablation} presents the experimental results of these schemes, indicating that both the emotion bank learning and the emotion discrimination objective significantly contribute to emotional expression and identity preservation, as reflected in the optimization of the F-SIM and Emo-Score metrics.

\noindent\textbf{Disentangled Emotion Embedder.}~
Table~\ref{tab:cluster_emo} provides a quantitative evaluation of the clustering strength of emotions after incorporating audio features. We define clustering strength $d_{cls} = \frac{d_{inter}}{d_{intra}}$ as the ratio between inter-cluster distance $d_{inter}$ and intra-cluster distance $d_{intra}$. A larger ratio indicates better clustering performance. We calculated the clustering strength of the emotion prior obtained under three conditions: using only video features, using only audio features, and using both together, with different emotions of the same identity as categories. The results show that visual information plays a crucial role in the extraction of emotion prior, while audio features can serve as an auxiliary to enhance the clustering strength of emotion prior. However, using only audio features is insufficient to obtain effective emotions.

When training the diffusion model in Sec.~\ref{diff_adapt}, if we do not employ an emotion prior sampling from Gaussian distribution in Eq.~\ref{eq:emotion} and instead use a deterministic mean $\mu_s$ for training, it may lead to overfitting of the training results and lost identity information.
Figure~\ref{fig:ablation_all}(b) shows the results of different emotion prior acquisition methods. Using a deterministic prior causes the model to transfer (eye reflections/facial contours) from the emotion reference video to the new face, leading to issues with identity deviation. The results in Table~\ref{tab:ablation} confirm this. By employing the disentangled emotion embedder, the emotion prior obtained by the model from the same training video varies each time, thereby preventing the transfer of incorrect content to the generated video.

\noindent\textbf{What Can We Learn from Emotion Prior?}~
We use t-SNE~\cite{van2008tsne} to project the emotion priors into a two-dimensional space. Figure~\ref{fig:ablation_all}(c) shows the emotion priors of two speakers from the MEAD dataset. Each prior is color-coded according to its corresponding emotion and intensity. The emotion priors with the same emotion first cluster together. Within each cluster, the emotion priors with the same intensity are closer to each other and there are noticeable transitions between emotion priors of different intensities. These observations indicate that our model can learn a continuous distribution of emotion embedding. As shown in Figure~\ref{fig:ablation_all}(d), when performing linear interpolation between two emotion embeddings extracted from the test set, the facial expressions and details in the generated video transition smoothly.
\section{Conclusion}
In summary, we propose DICE-Talk, a method for generating talking head videos with emotion control. By designing and training the disentangled emotion embedder, we obtain emotion priors that can fully represent the emotions and habits of the emotion reference video. Through the correlation-enhanced emotion conditioning module, we further capture the relationships between emotions, thus better representing various emotions. Finally, with the design of the compositional diffusion adaptation, we successfully transfer emotion priors to unseen faces. Experimental results show that DICE-Talk not only achieves vivid and rich emotional expressions but also ensures the overall quality of the generated video, providing new insights for more advanced and comprehensive talking head video generation.

\bibliographystyle{ACM-Reference-Format}
\balance
\bibliography{acmart}


\begin{thebibliography}{47}


\ifx \showCODEN    \undefined \def \showCODEN     #1{\unskip}     \fi
\ifx \showISBNx    \undefined \def \showISBNx     #1{\unskip}     \fi
\ifx \showISBNxiii \undefined \def \showISBNxiii  #1{\unskip}     \fi
\ifx \showISSN     \undefined \def \showISSN      #1{\unskip}     \fi
\ifx \showLCCN     \undefined \def \showLCCN      #1{\unskip}     \fi
\ifx \shownote     \undefined \def \shownote      #1{#1}          \fi
\ifx \showarticletitle \undefined \def \showarticletitle #1{#1}   \fi
\ifx \showURL      \undefined \def \showURL       {\relax}        \fi
\providecommand\bibfield[2]{#2}
\providecommand\bibinfo[2]{#2}
\providecommand\natexlab[1]{#1}
\providecommand\showeprint[2][]{arXiv:#2}

\bibitem[Blanz and Vetter(1999)]%
        {blanz1999morphable}
\bibfield{author}{\bibinfo{person}{Volker Blanz} {and} \bibinfo{person}{Thomas Vetter}.} \bibinfo{year}{1999}\natexlab{}.
\newblock \showarticletitle{A morphable model for the synthesis of 3D faces}. In \bibinfo{booktitle}{\emph{ACM SIGGRAPH}}.
\newblock


\bibitem[Blattmann et~al\mbox{.}(2023a)]%
        {blattmann2023svd}
\bibfield{author}{\bibinfo{person}{Andreas Blattmann}, \bibinfo{person}{Tim Dockhorn}, \bibinfo{person}{Sumith Kulal}, \bibinfo{person}{Daniel Mendelevitch}, \bibinfo{person}{Maciej Kilian}, \bibinfo{person}{Dominik Lorenz}, \bibinfo{person}{Yam Levi}, \bibinfo{person}{Zion English}, \bibinfo{person}{Vikram Voleti}, \bibinfo{person}{Adam Letts}, {et~al\mbox{.}}} \bibinfo{year}{2023}\natexlab{a}.
\newblock \showarticletitle{Stable video diffusion: Scaling latent video diffusion models to large datasets}.
\newblock \bibinfo{journal}{\emph{arXiv preprint arXiv:2311.15127}} (\bibinfo{year}{2023}).
\newblock


\bibitem[Blattmann et~al\mbox{.}(2023b)]%
        {blattmann2023stable}
\bibfield{author}{\bibinfo{person}{Andreas Blattmann}, \bibinfo{person}{Tim Dockhorn}, \bibinfo{person}{Sumith Kulal}, \bibinfo{person}{Daniel Mendelevitch}, \bibinfo{person}{Maciej Kilian}, \bibinfo{person}{Dominik Lorenz}, \bibinfo{person}{Yam Levi}, \bibinfo{person}{Zion English}, \bibinfo{person}{Vikram Voleti}, \bibinfo{person}{Adam Letts}, {et~al\mbox{.}}} \bibinfo{year}{2023}\natexlab{b}.
\newblock \showarticletitle{Stable video diffusion: Scaling latent video diffusion models to large datasets}.
\newblock \bibinfo{journal}{\emph{arXiv preprint arXiv:2311.15127}} (\bibinfo{year}{2023}).
\newblock


\bibitem[Chen et~al\mbox{.}(2020)]%
        {chen2020simclr}
\bibfield{author}{\bibinfo{person}{Ting Chen}, \bibinfo{person}{Simon Kornblith}, \bibinfo{person}{Mohammad Norouzi}, {and} \bibinfo{person}{Geoffrey Hinton}.} \bibinfo{year}{2020}\natexlab{}.
\newblock \showarticletitle{A simple framework for contrastive learning of visual representations}. In \bibinfo{booktitle}{\emph{ICML}}.
\newblock


\bibitem[Chen et~al\mbox{.}(2024)]%
        {chen2024echomimic}
\bibfield{author}{\bibinfo{person}{Zhiyuan Chen}, \bibinfo{person}{Jiajiong Cao}, \bibinfo{person}{Zhiquan Chen}, \bibinfo{person}{Yuming Li}, {and} \bibinfo{person}{Chenguang Ma}.} \bibinfo{year}{2024}\natexlab{}.
\newblock \showarticletitle{Echomimic: Lifelike audio-driven portrait animations through editable landmark conditions}.
\newblock \bibinfo{journal}{\emph{arXiv preprint arXiv:2407.08136}} (\bibinfo{year}{2024}).
\newblock


\bibitem[Cheng et~al\mbox{.}(2022)]%
        {cheng2022videoretalking}
\bibfield{author}{\bibinfo{person}{Kun Cheng}, \bibinfo{person}{Xiaodong Cun}, \bibinfo{person}{Yong Zhang}, \bibinfo{person}{Menghan Xia}, \bibinfo{person}{Fei Yin}, \bibinfo{person}{Mingrui Zhu}, \bibinfo{person}{Xuan Wang}, \bibinfo{person}{Jue Wang}, {and} \bibinfo{person}{Nannan Wang}.} \bibinfo{year}{2022}\natexlab{}.
\newblock \showarticletitle{Videoretalking: Audio-based lip synchronization for talking head video editing in the wild}. In \bibinfo{booktitle}{\emph{ACM SIGGRAPH Asia}}.
\newblock


\bibitem[Chung et~al\mbox{.}(2018)]%
        {Chung_2018}
\bibfield{author}{\bibinfo{person}{Joon~Son Chung}, \bibinfo{person}{Arsha Nagrani}, {and} \bibinfo{person}{Andrew Zisserman}.} \bibinfo{year}{2018}\natexlab{}.
\newblock \showarticletitle{VoxCeleb2: Deep Speaker Recognition}. In \bibinfo{booktitle}{\emph{Interspeech 2018}}. \bibinfo{publisher}{ISCA}.
\newblock
\href{https://doi.org/10.21437/interspeech.2018-1929}{doi:\nolinkurl{10.21437/interspeech.2018-1929}}


\bibitem[Chung and Zisserman(2017)]%
        {chung2017syncnet}
\bibfield{author}{\bibinfo{person}{Joon~Son Chung} {and} \bibinfo{person}{Andrew Zisserman}.} \bibinfo{year}{2017}\natexlab{}.
\newblock \showarticletitle{Out of time: automated lip sync in the wild}. In \bibinfo{booktitle}{\emph{ACCV Workshops}}.
\newblock


\bibitem[Cui et~al\mbox{.}(2024)]%
        {cui2024hallo2}
\bibfield{author}{\bibinfo{person}{Jiahao Cui}, \bibinfo{person}{Hui Li}, \bibinfo{person}{Yao Yao}, \bibinfo{person}{Hao Zhu}, \bibinfo{person}{Hanlin Shang}, \bibinfo{person}{Kaihui Cheng}, \bibinfo{person}{Hang Zhou}, \bibinfo{person}{Siyu Zhu}, {and} \bibinfo{person}{Jingdong Wang}.} \bibinfo{year}{2024}\natexlab{}.
\newblock \showarticletitle{Hallo2: Long-duration and high-resolution audio-driven portrait image animation}.
\newblock \bibinfo{journal}{\emph{arXiv preprint arXiv:2410.07718}} (\bibinfo{year}{2024}).
\newblock


\bibitem[Gan et~al\mbox{.}(2023)]%
        {gan2023eat}
\bibfield{author}{\bibinfo{person}{Yuan Gan}, \bibinfo{person}{Zongxin Yang}, \bibinfo{person}{Xihang Yue}, \bibinfo{person}{Lingyun Sun}, {and} \bibinfo{person}{Yi Yang}.} \bibinfo{year}{2023}\natexlab{}.
\newblock \showarticletitle{Efficient emotional adaptation for audio-driven talking-head generation}. In \bibinfo{booktitle}{\emph{Proceedings of the IEEE/CVF International Conference on Computer Vision}}. \bibinfo{pages}{22634--22645}.
\newblock


\bibitem[Heusel et~al\mbox{.}(2017)]%
        {heusel2017fid}
\bibfield{author}{\bibinfo{person}{Martin Heusel}, \bibinfo{person}{Hubert Ramsauer}, \bibinfo{person}{Thomas Unterthiner}, \bibinfo{person}{Bernhard Nessler}, {and} \bibinfo{person}{Sepp Hochreiter}.} \bibinfo{year}{2017}\natexlab{}.
\newblock \showarticletitle{Gans trained by a two time-scale update rule converge to a local nash equilibrium}. In \bibinfo{booktitle}{\emph{NeurIPS}}.
\newblock


\bibitem[Huang et~al\mbox{.}(2024)]%
        {huang2024learning}
\bibfield{author}{\bibinfo{person}{Siteng Huang}, \bibinfo{person}{Biao Gong}, \bibinfo{person}{Yutong Feng}, \bibinfo{person}{Xi Chen}, \bibinfo{person}{Yuqian Fu}, \bibinfo{person}{Yu Liu}, {and} \bibinfo{person}{Donglin Wang}.} \bibinfo{year}{2024}\natexlab{}.
\newblock \showarticletitle{Learning disentangled identifiers for action-customized text-to-image generation}. In \bibinfo{booktitle}{\emph{Proceedings of the IEEE/CVF Conference on Computer Vision and Pattern Recognition}}. \bibinfo{pages}{7797--7806}.
\newblock


\bibitem[Ji et~al\mbox{.}(2024)]%
        {ji2024sonic}
\bibfield{author}{\bibinfo{person}{Xiaozhong Ji}, \bibinfo{person}{Xiaobin Hu}, \bibinfo{person}{Zhihong Xu}, \bibinfo{person}{Junwei Zhu}, \bibinfo{person}{Chuming Lin}, \bibinfo{person}{Qingdong He}, \bibinfo{person}{Jiangning Zhang}, \bibinfo{person}{Donghao Luo}, \bibinfo{person}{Yi Chen}, \bibinfo{person}{Qin Lin}, {et~al\mbox{.}}} \bibinfo{year}{2024}\natexlab{}.
\newblock \showarticletitle{Sonic: Shifting Focus to Global Audio Perception in Portrait Animation}.
\newblock \bibinfo{journal}{\emph{arXiv preprint arXiv:2411.16331}} (\bibinfo{year}{2024}).
\newblock


\bibitem[Ji et~al\mbox{.}(2022)]%
        {ji2022eamm}
\bibfield{author}{\bibinfo{person}{Xinya Ji}, \bibinfo{person}{Hang Zhou}, \bibinfo{person}{Kaisiyuan Wang}, \bibinfo{person}{Qianyi Wu}, \bibinfo{person}{Wayne Wu}, \bibinfo{person}{Feng Xu}, {and} \bibinfo{person}{Xun Cao}.} \bibinfo{year}{2022}\natexlab{}.
\newblock \showarticletitle{Eamm: One-shot emotional talking face via audio-based emotion-aware motion model}. In \bibinfo{booktitle}{\emph{ACM SIGGRAPH}}.
\newblock


\bibitem[Ji et~al\mbox{.}(2021)]%
        {ji2021evp}
\bibfield{author}{\bibinfo{person}{Xinya Ji}, \bibinfo{person}{Hang Zhou}, \bibinfo{person}{Kaisiyuan Wang}, \bibinfo{person}{Wayne Wu}, \bibinfo{person}{Chen~Change Loy}, \bibinfo{person}{Xun Cao}, {and} \bibinfo{person}{Feng Xu}.} \bibinfo{year}{2021}\natexlab{}.
\newblock \showarticletitle{Audio-driven emotional video portraits}. In \bibinfo{booktitle}{\emph{CVPR}}.
\newblock


\bibitem[Jiang et~al\mbox{.}(2024)]%
        {jiang2024loopy}
\bibfield{author}{\bibinfo{person}{Jianwen Jiang}, \bibinfo{person}{Chao Liang}, \bibinfo{person}{Jiaqi Yang}, \bibinfo{person}{Gaojie Lin}, \bibinfo{person}{Tianyun Zhong}, {and} \bibinfo{person}{Yanbo Zheng}.} \bibinfo{year}{2024}\natexlab{}.
\newblock \showarticletitle{Loopy: Taming audio-driven portrait avatar with long-term motion dependency}.
\newblock \bibinfo{journal}{\emph{arXiv preprint arXiv:2409.02634}} (\bibinfo{year}{2024}).
\newblock


\bibitem[Karras et~al\mbox{.}(2022)]%
        {karras2022edm}
\bibfield{author}{\bibinfo{person}{Tero Karras}, \bibinfo{person}{Miika Aittala}, \bibinfo{person}{Timo Aila}, {and} \bibinfo{person}{Samuli Laine}.} \bibinfo{year}{2022}\natexlab{}.
\newblock \showarticletitle{Elucidating the design space of diffusion-based generative models}.
\newblock \bibinfo{journal}{\emph{Advances in neural information processing systems}}  \bibinfo{volume}{35} (\bibinfo{year}{2022}), \bibinfo{pages}{26565--26577}.
\newblock


\bibitem[Kingma and Ba(2015)]%
        {kingma2014adam}
\bibfield{author}{\bibinfo{person}{Diederik~P Kingma} {and} \bibinfo{person}{Jimmy Ba}.} \bibinfo{year}{2015}\natexlab{}.
\newblock \showarticletitle{Adam: A method for stochastic optimization}. In \bibinfo{booktitle}{\emph{ICLR}}.
\newblock


\bibitem[Liang et~al\mbox{.}(2022)]%
        {liang2022expressive}
\bibfield{author}{\bibinfo{person}{Borong Liang}, \bibinfo{person}{Yan Pan}, \bibinfo{person}{Zhizhi Guo}, \bibinfo{person}{Hang Zhou}, \bibinfo{person}{Zhibin Hong}, \bibinfo{person}{Xiaoguang Han}, \bibinfo{person}{Junyu Han}, \bibinfo{person}{Jingtuo Liu}, \bibinfo{person}{Errui Ding}, {and} \bibinfo{person}{Jingdong Wang}.} \bibinfo{year}{2022}\natexlab{}.
\newblock \showarticletitle{Expressive talking head generation with granular audio-visual control}. In \bibinfo{booktitle}{\emph{CVPR}}.
\newblock


\bibitem[Ma et~al\mbox{.}(2023)]%
        {ma2023styletalk}
\bibfield{author}{\bibinfo{person}{Yifeng Ma}, \bibinfo{person}{Suzhen Wang}, \bibinfo{person}{Zhipeng Hu}, \bibinfo{person}{Changjie Fan}, \bibinfo{person}{Tangjie Lv}, \bibinfo{person}{Yu Ding}, \bibinfo{person}{Zhidong Deng}, {and} \bibinfo{person}{Xin Yu}.} \bibinfo{year}{2023}\natexlab{}.
\newblock \showarticletitle{Styletalk: One-shot talking head generation with controllable speaking styles}. In \bibinfo{booktitle}{\emph{AAAI}}.
\newblock


\bibitem[Pan et~al\mbox{.}(2025)]%
        {pan2025earthsynth}
\bibfield{author}{\bibinfo{person}{Jiancheng Pan}, \bibinfo{person}{Shiye Lei}, \bibinfo{person}{Yuqian Fu}, \bibinfo{person}{Jiahao Li}, \bibinfo{person}{Yanxing Liu}, \bibinfo{person}{Yuze Sun}, \bibinfo{person}{Xiao He}, \bibinfo{person}{Long Peng}, \bibinfo{person}{Xiaomeng Huang}, {and} \bibinfo{person}{Bo Zhao}.} \bibinfo{year}{2025}\natexlab{}.
\newblock \showarticletitle{EarthSynth: Generating Informative Earth Observation with Diffusion Models}.
\newblock \bibinfo{journal}{\emph{arXiv preprint arXiv:2505.12108}} (\bibinfo{year}{2025}).
\newblock


\bibitem[Paszke et~al\mbox{.}(2019)]%
        {paszke2019pytorch}
\bibfield{author}{\bibinfo{person}{Adam Paszke}, \bibinfo{person}{Sam Gross}, \bibinfo{person}{Francisco Massa}, \bibinfo{person}{Adam Lerer}, \bibinfo{person}{James Bradbury}, \bibinfo{person}{Gregory Chanan}, \bibinfo{person}{Trevor Killeen}, \bibinfo{person}{Zeming Lin}, \bibinfo{person}{Natalia Gimelshein}, \bibinfo{person}{Luca Antiga}, {et~al\mbox{.}}} \bibinfo{year}{2019}\natexlab{}.
\newblock \showarticletitle{Pytorch: An imperative style, high-performance deep learning library}. In \bibinfo{booktitle}{\emph{NeurIPS}}.
\newblock


\bibitem[Prajwal et~al\mbox{.}(2020)]%
        {prajwal2020wav2lip}
\bibfield{author}{\bibinfo{person}{KR Prajwal}, \bibinfo{person}{Rudrabha Mukhopadhyay}, \bibinfo{person}{Vinay~P Namboodiri}, {and} \bibinfo{person}{CV Jawahar}.} \bibinfo{year}{2020}\natexlab{}.
\newblock \showarticletitle{A lip sync expert is all you need for speech to lip generation in the wild}. In \bibinfo{booktitle}{\emph{ACM MM}}.
\newblock


\bibitem[Radford et~al\mbox{.}(2023)]%
        {radford2023whisper}
\bibfield{author}{\bibinfo{person}{Alec Radford}, \bibinfo{person}{Jong~Wook Kim}, \bibinfo{person}{Tao Xu}, \bibinfo{person}{Greg Brockman}, \bibinfo{person}{Christine McLeavey}, {and} \bibinfo{person}{Ilya Sutskever}.} \bibinfo{year}{2023}\natexlab{}.
\newblock \showarticletitle{Robust speech recognition via large-scale weak supervision}. In \bibinfo{booktitle}{\emph{ICML}}.
\newblock


\bibitem[Rombach et~al\mbox{.}(2022)]%
        {rombach2022ldm}
\bibfield{author}{\bibinfo{person}{Robin Rombach}, \bibinfo{person}{Andreas Blattmann}, \bibinfo{person}{Dominik Lorenz}, \bibinfo{person}{Patrick Esser}, {and} \bibinfo{person}{Bj{\"o}rn Ommer}.} \bibinfo{year}{2022}\natexlab{}.
\newblock \showarticletitle{High-resolution image synthesis with latent diffusion models}. In \bibinfo{booktitle}{\emph{CVPR}}.
\newblock


\bibitem[Ryumina et~al\mbox{.}(2022)]%
        {RYUMINA2022emoscore}
\bibfield{author}{\bibinfo{person}{Elena Ryumina}, \bibinfo{person}{Denis Dresvyanskiy}, {and} \bibinfo{person}{Alexey Karpov}.} \bibinfo{year}{2022}\natexlab{}.
\newblock \showarticletitle{In Search of a Robust Facial Expressions Recognition Model: A Large-Scale Visual Cross-Corpus Study}.
\newblock \bibinfo{journal}{\emph{Neurocomputing}} (\bibinfo{year}{2022}).
\newblock
\href{https://doi.org/10.1016/j.neucom.2022.10.013}{doi:\nolinkurl{10.1016/j.neucom.2022.10.013}}


\bibitem[Shen et~al\mbox{.}(2023)]%
        {shen2023difftalk}
\bibfield{author}{\bibinfo{person}{Shuai Shen}, \bibinfo{person}{Wenliang Zhao}, \bibinfo{person}{Zibin Meng}, \bibinfo{person}{Wanhua Li}, \bibinfo{person}{Zheng Zhu}, \bibinfo{person}{Jie Zhou}, {and} \bibinfo{person}{Jiwen Lu}.} \bibinfo{year}{2023}\natexlab{}.
\newblock \showarticletitle{Difftalk: Crafting diffusion models for generalized audio-driven portraits animation}. In \bibinfo{booktitle}{\emph{CVPR}}.
\newblock


\bibitem[Sinha et~al\mbox{.}(2022)]%
        {sinha2022emotion}
\bibfield{author}{\bibinfo{person}{Sanjana Sinha}, \bibinfo{person}{Sandika Biswas}, \bibinfo{person}{Ravindra Yadav}, {and} \bibinfo{person}{Brojeshwar Bhowmick}.} \bibinfo{year}{2022}\natexlab{}.
\newblock \showarticletitle{Emotion-controllable generalized talking face generation}. In \bibinfo{booktitle}{\emph{IJCAI}}.
\newblock


\bibitem[Tian et~al\mbox{.}(2024)]%
        {tian2024emo}
\bibfield{author}{\bibinfo{person}{Linrui Tian}, \bibinfo{person}{Qi Wang}, \bibinfo{person}{Bang Zhang}, {and} \bibinfo{person}{Liefeng Bo}.} \bibinfo{year}{2024}\natexlab{}.
\newblock \showarticletitle{Emo: Emote portrait alive-generating expressive portrait videos with audio2video diffusion model under weak conditions}. In \bibinfo{booktitle}{\emph{ECCV}}.
\newblock


\bibitem[Unterthiner et~al\mbox{.}(2019)]%
        {unterthiner2018fvd}
\bibfield{author}{\bibinfo{person}{Thomas Unterthiner}, \bibinfo{person}{Sjoerd van Steenkiste}, \bibinfo{person}{Karol Kurach}, \bibinfo{person}{Rapha{\"e}l Marinier}, \bibinfo{person}{Marcin Michalski}, {and} \bibinfo{person}{Sylvain Gelly}.} \bibinfo{year}{2019}\natexlab{}.
\newblock \showarticletitle{FVD: A new metric for video generation}. In \bibinfo{booktitle}{\emph{ICLR Workshops}}.
\newblock


\bibitem[Van Den~Oord et~al\mbox{.}(2017)]%
        {van2017vqvae}
\bibfield{author}{\bibinfo{person}{Aaron Van Den~Oord}, \bibinfo{person}{Oriol Vinyals}, {et~al\mbox{.}}} \bibinfo{year}{2017}\natexlab{}.
\newblock \showarticletitle{Neural discrete representation learning}.
\newblock \bibinfo{journal}{\emph{Advances in neural information processing systems}}  \bibinfo{volume}{30} (\bibinfo{year}{2017}).
\newblock


\bibitem[Van~der Maaten and Hinton(2008)]%
        {van2008tsne}
\bibfield{author}{\bibinfo{person}{Laurens Van~der Maaten} {and} \bibinfo{person}{Geoffrey Hinton}.} \bibinfo{year}{2008}\natexlab{}.
\newblock \showarticletitle{Visualizing data using t-SNE.}
\newblock \bibinfo{journal}{\emph{Journal of machine learning research}} (\bibinfo{year}{2008}).
\newblock


\bibitem[Wang et~al\mbox{.}(2024)]%
        {wang2024vexpress}
\bibfield{author}{\bibinfo{person}{Cong Wang}, \bibinfo{person}{Kuan Tian}, \bibinfo{person}{Jun Zhang}, \bibinfo{person}{Yonghang Guan}, \bibinfo{person}{Feng Luo}, \bibinfo{person}{Fei Shen}, \bibinfo{person}{Zhiwei Jiang}, \bibinfo{person}{Qing Gu}, \bibinfo{person}{Xiao Han}, {and} \bibinfo{person}{Wei Yang}.} \bibinfo{year}{2024}\natexlab{}.
\newblock \showarticletitle{V-Express: Conditional Dropout for Progressive Training of Portrait Video Generation}.
\newblock \bibinfo{journal}{\emph{arXiv preprint arXiv:2406.02511}} (\bibinfo{year}{2024}).
\newblock


\bibitem[Wang et~al\mbox{.}(2023)]%
        {wang2023lipread}
\bibfield{author}{\bibinfo{person}{Jiadong Wang}, \bibinfo{person}{Xinyuan Qian}, \bibinfo{person}{Malu Zhang}, \bibinfo{person}{Robby~T Tan}, {and} \bibinfo{person}{Haizhou Li}.} \bibinfo{year}{2023}\natexlab{}.
\newblock \showarticletitle{Seeing What You Said: Talking Face Generation Guided by a Lip Reading Expert}. In \bibinfo{booktitle}{\emph{CVPR}}.
\newblock


\bibitem[Wang et~al\mbox{.}(2020)]%
        {wang2020mead}
\bibfield{author}{\bibinfo{person}{Kaisiyuan Wang}, \bibinfo{person}{Qianyi Wu}, \bibinfo{person}{Linsen Song}, \bibinfo{person}{Zhuoqian Yang}, \bibinfo{person}{Wayne Wu}, \bibinfo{person}{Chen Qian}, \bibinfo{person}{Ran He}, \bibinfo{person}{Yu Qiao}, {and} \bibinfo{person}{Chen~Change Loy}.} \bibinfo{year}{2020}\natexlab{}.
\newblock \showarticletitle{Mead: A large-scale audio-visual dataset for emotional talking-face generation}. In \bibinfo{booktitle}{\emph{ECCV}}.
\newblock


\bibitem[Wang et~al\mbox{.}(2021)]%
        {wang2021audio2head}
\bibfield{author}{\bibinfo{person}{Suzhen Wang}, \bibinfo{person}{Lincheng Li}, \bibinfo{person}{Yu Ding}, \bibinfo{person}{Changjie Fan}, {and} \bibinfo{person}{Xin Yu}.} \bibinfo{year}{2021}\natexlab{}.
\newblock \showarticletitle{Audio2head: Audio-driven one-shot talking-head generation with natural head motion}. In \bibinfo{booktitle}{\emph{IJCAI}}.
\newblock


\bibitem[Wang et~al\mbox{.}(2022)]%
        {wang2022one}
\bibfield{author}{\bibinfo{person}{Suzhen Wang}, \bibinfo{person}{Lincheng Li}, \bibinfo{person}{Yu Ding}, {and} \bibinfo{person}{Xin Yu}.} \bibinfo{year}{2022}\natexlab{}.
\newblock \showarticletitle{One-shot talking face generation from single-speaker audio-visual correlation learning}. In \bibinfo{booktitle}{\emph{AAAI}}.
\newblock


\bibitem[Wei et~al\mbox{.}(2024)]%
        {wei2024aniportrait}
\bibfield{author}{\bibinfo{person}{Huawei Wei}, \bibinfo{person}{Zejun Yang}, {and} \bibinfo{person}{Zhisheng Wang}.} \bibinfo{year}{2024}\natexlab{}.
\newblock \showarticletitle{Aniportrait: Audio-driven synthesis of photorealistic portrait animation}.
\newblock \bibinfo{journal}{\emph{arXiv preprint arXiv:2403.17694}} (\bibinfo{year}{2024}).
\newblock


\bibitem[Wu et~al\mbox{.}(2021)]%
        {wu2021imitating}
\bibfield{author}{\bibinfo{person}{Haozhe Wu}, \bibinfo{person}{Jia Jia}, \bibinfo{person}{Haoyu Wang}, \bibinfo{person}{Yishun Dou}, \bibinfo{person}{Chao Duan}, {and} \bibinfo{person}{Qingshan Deng}.} \bibinfo{year}{2021}\natexlab{}.
\newblock \showarticletitle{Imitating arbitrary talking style for realistic audio-driven talking face synthesis}. In \bibinfo{booktitle}{\emph{ACM MM}}.
\newblock


\bibitem[Wu et~al\mbox{.}(2023)]%
        {wu2023tune}
\bibfield{author}{\bibinfo{person}{Jay~Zhangjie Wu}, \bibinfo{person}{Yixiao Ge}, \bibinfo{person}{Xintao Wang}, \bibinfo{person}{Stan~Weixian Lei}, \bibinfo{person}{Yuchao Gu}, \bibinfo{person}{Yufei Shi}, \bibinfo{person}{Wynne Hsu}, \bibinfo{person}{Ying Shan}, \bibinfo{person}{Xiaohu Qie}, {and} \bibinfo{person}{Mike~Zheng Shou}.} \bibinfo{year}{2023}\natexlab{}.
\newblock \showarticletitle{Tune-a-video: One-shot tuning of image diffusion models for text-to-video generation}. In \bibinfo{booktitle}{\emph{ICCV}}.
\newblock


\bibitem[Xie et~al\mbox{.}(2024)]%
        {xie2024xportrait}
\bibfield{author}{\bibinfo{person}{You Xie}, \bibinfo{person}{Hongyi Xu}, \bibinfo{person}{Guoxian Song}, \bibinfo{person}{Chao Wang}, \bibinfo{person}{Yichun Shi}, {and} \bibinfo{person}{Linjie Luo}.} \bibinfo{year}{2024}\natexlab{}.
\newblock \showarticletitle{X-portrait: Expressive portrait animation with hierarchical motion attention}. In \bibinfo{booktitle}{\emph{ACM SIGGRAPH}}.
\newblock


\bibitem[Xu et~al\mbox{.}(2024)]%
        {xu2024hallo}
\bibfield{author}{\bibinfo{person}{Mingwang Xu}, \bibinfo{person}{Hui Li}, \bibinfo{person}{Qingkun Su}, \bibinfo{person}{Hanlin Shang}, \bibinfo{person}{Liwei Zhang}, \bibinfo{person}{Ce Liu}, \bibinfo{person}{Jingdong Wang}, \bibinfo{person}{Luc Van~Gool}, \bibinfo{person}{Yao Yao}, {and} \bibinfo{person}{Siyu Zhu}.} \bibinfo{year}{2024}\natexlab{}.
\newblock \showarticletitle{Hallo: Hierarchical Audio-Driven Visual Synthesis for Portrait Image Animation}.
\newblock \bibinfo{journal}{\emph{arXiv preprint arXiv:2406.08801}} (\bibinfo{year}{2024}).
\newblock


\bibitem[Yang et~al\mbox{.}(2024)]%
        {yang2024megactor}
\bibfield{author}{\bibinfo{person}{Shurong Yang}, \bibinfo{person}{Huadong Li}, \bibinfo{person}{Juhao Wu}, \bibinfo{person}{Minhao Jing}, \bibinfo{person}{Linze Li}, \bibinfo{person}{Renhe Ji}, \bibinfo{person}{Jiajun Liang}, {and} \bibinfo{person}{Haoqiang Fan}.} \bibinfo{year}{2024}\natexlab{}.
\newblock \showarticletitle{MegActor: Harness the Power of Raw Video for Vivid Portrait Animation}.
\newblock \bibinfo{journal}{\emph{arXiv preprint arXiv:2405.20851}} (\bibinfo{year}{2024}).
\newblock


\bibitem[Zhang et~al\mbox{.}(2023)]%
        {zhang2023sadtalker}
\bibfield{author}{\bibinfo{person}{Wenxuan Zhang}, \bibinfo{person}{Xiaodong Cun}, \bibinfo{person}{Xuan Wang}, \bibinfo{person}{Yong Zhang}, \bibinfo{person}{Xi Shen}, \bibinfo{person}{Yu Guo}, \bibinfo{person}{Ying Shan}, {and} \bibinfo{person}{Fei Wang}.} \bibinfo{year}{2023}\natexlab{}.
\newblock \showarticletitle{Sadtalker: Learning realistic 3d motion coefficients for stylized audio-driven single image talking face animation}. In \bibinfo{booktitle}{\emph{CVPR}}.
\newblock


\bibitem[Zhang et~al\mbox{.}(2021)]%
        {zhang2021hdtf}
\bibfield{author}{\bibinfo{person}{Zhimeng Zhang}, \bibinfo{person}{Lincheng Li}, \bibinfo{person}{Yu Ding}, {and} \bibinfo{person}{Changjie Fan}.} \bibinfo{year}{2021}\natexlab{}.
\newblock \showarticletitle{Flow-guided one-shot talking face generation with a high-resolution audio-visual dataset}. In \bibinfo{booktitle}{\emph{CVPR}}.
\newblock


\bibitem[Zhou et~al\mbox{.}(2021)]%
        {zhou2021pcavs}
\bibfield{author}{\bibinfo{person}{Hang Zhou}, \bibinfo{person}{Yasheng Sun}, \bibinfo{person}{Wayne Wu}, \bibinfo{person}{Chen~Change Loy}, \bibinfo{person}{Xiaogang Wang}, {and} \bibinfo{person}{Ziwei Liu}.} \bibinfo{year}{2021}\natexlab{}.
\newblock \showarticletitle{Pose-controllable talking face generation by implicitly modularized audio-visual representation}. In \bibinfo{booktitle}{\emph{CVPR}}.
\newblock


\bibitem[Zhou et~al\mbox{.}(2020)]%
        {zhou2020makeittalk}
\bibfield{author}{\bibinfo{person}{Yang Zhou}, \bibinfo{person}{Xintong Han}, \bibinfo{person}{Eli Shechtman}, \bibinfo{person}{Jose Echevarria}, \bibinfo{person}{Evangelos Kalogerakis}, {and} \bibinfo{person}{Dingzeyu Li}.} \bibinfo{year}{2020}\natexlab{}.
\newblock \showarticletitle{Makelttalk: speaker-aware talking-head animation}.
\newblock \bibinfo{journal}{\emph{ACM TOG}} (\bibinfo{year}{2020}).
\newblock


\end{thebibliography}










\end{document}